# Application of Fuzzy Mathematics to Speech-to-Text Conversion by Elimination of Paralinguistic Content

Sachin Lakra*, T.V. Prasad**, Deepak Kumar Sharma*, Shree Harsh Atrey*,
Anubhav Kumar Sharma**

*Manav Rachna College of Engineering, Faridabad, Haryana, India.
**Lingaya's Institute of Management and Technology, Faridabad, Haryana, India.

*Abstract -* **For the past few decades, man has been trying to create an intelligent computer which can talk and respond like he can. The task of creating a system that can talk like a human being is the primary objective of Automatic Speech Recognition. Various Speech Recognition techniques have been developed in theory and have been applied in practice. This paper discusses the problems that have been encountered in developing Speech Recognition, the techniques that have been applied to automate the task, and a representation of the core problems of present day Speech Recognition by using Fuzzy Mathematics.**

*Keywords*: *Fuzzy Mathematics, Automatic Speech Recognition, Dynamic Time Warping, Hidden Markov Models, Hybrid Neural Networks, Neuro-Fuzzy Systems.*

1. INTRODUCTION

For ages, man has tried to create systems that can think, work and talk like he can. In this pursuit, man has created machines, the computer and software systems that are almost intelligent. The ultimate objective of solving these problems is to create an intelligent system that is humanoid in appearance and which can live the life of a human being, although as a machine. A part of this endeavour is the development of a system that listens and responds verbally, just like human beings can. The means of achieving this goal is to automate speech recognition tasks and man has achieved significant success, although not perfection, in this direction, till date.

Section 1 introduces the paper and Section 2 gives a definition of Automatic Speech Recognition. Section 3 briefly presents the problems that have been encountered in developing Automatic Speech Recognition. Section 3 further describes the techniques that have been developed over the past few decades to solve these problems and identifies the problems that are the core of a major number of the problems in this area. Section 4 presents a representation of these core problems, using Fuzzy Mathematics.

2. SPEECH RECOGNITION

Speech recognition, or more commonly known as Automatic Speech Recognition (ASR), is the process of interpreting human speech by a computer[7]. A more technical definition is given by Jurafsky[2], where he defines ASR as the building of a system for mapping acoustic signals to a string of words [2]. Speech has two types of content, namely, linguistic content and paralinguistic content. Linguistic content refers to that content which is part of the language in the written form, for example, grammar, semantics, etc. Paralinguistic content refers to the content other than linguistic content, which is part of the spoken content, for example, emphasis, accent, gestures,etc.

3. PROBLEMS IN SPEECH RECOGNITION AND TECHNIQUES USED

*3.1.    Problems in Speech Recognition*
The following paragraphs present the problems which have been encountered in developing ASR.

*3.1.1.    Human comprehension of speech compared to ASR*
Humans use more than their ears when listening. They use the knowledge they have about the speaker and the subject. Words are not arbitrarily sequenced together, there is a grammatical structure and redundancy that humans use to predict words not yet spoken. Furthermore, idioms and how people 'usually' say things makes prediction even easier.

In ASR there is only the speech signal. A model of the grammatical structure can be constructed and some kind of statistical model can be used to improve prediction [7].

*3.1.2 Body language*
A human speaker does not only communicate with speech, but also with body signals - hand waving, eye movements, postures, etc. This information is completely missed by ASR [7].

*3.1.3 Noise*
Speech is uttered in an environment of sounds, a clock ticking, a computer humming, a radio playing somewhere down the corridor, another human speaker in the background etc. This is usually called noise, i.e., unwanted information in the speech signal. In ASR this noise has to be identified and filtered out from the speech signal [7].

*3.1.4 Is Spoken language equal to Written language?*
Spoken language has for many years been viewed just as a less complicated version of written language, with the main difference that spoken language is grammatically less complex and that humans make more performance errors while speaking. However, it has become clear in the last few years that spoken language is essentially different from written language. In ASR, these differences have to be identified and addressed [7].

*3.1.5 Continuous speech*
Speech has no natural pauses between the word boundaries; the pauses mainly appear on a syntactic level, such as after a phrase or a sentence.

This introduces a difficult problem for speech recognition – how to translate a waveform into a sequence of words [7]?

*3.1.6 Channel variability*
One aspect of variability is the context where the acoustic wave is uttered. The problem here is that noise changes over time and different kinds of microphones affect the content of the acoustic wave from the speaker to the discrete representation in a computer. This phenomenon is called channel variability [7].

*3.1.7 Speaker variability*
All speakers have their special voices, due to their unique physical body and personality. The voice is not only different between speakers, there are also wide variations within one specific speaker.

Some of these variations include the following:

- *Realization*: If the same words were pronounced over and over again, the resulting speech signal would never look exactly the same. That is, the realization of speech changes over time.
- *Speaking style*: All humans speak differently as it is a way of expressing their personality. Not only do they use a personal vocabulary, but they have a unique way to pronounce and emphasize. The speaking style also varies in different situations, such as speaking in a bus, or speaking to a friend. Humans also communicate their emotions via speech, for example, while speaking in happiness, sadness, under stress, under disappointment, etc.
- *The gender of the speaker*: Men and women have different voices, and the main reason for this is that women have in general a shorter vocal tract than men. The fundamental tone of female voices is roughly two times higher than male voices because of this difference.
- *Anatomy of the vocal tract*: The shape and length of the vocal cords, the formation of the cavities, the size of the lungs, etc. are various attributes of the tract. These attributes change over time, e.g., depending on the health or the age of the speaker.
- *Speed of speech*: Humans speak at different speeds at different times. If a person is stressed, he tends to speak faster, and if he is tired, the speed tends to decrease.
- *Regional and social dialects*: Dialects are group related variations within a language. In many cases, it is better to consider dialects as 'another language' in ASR, due to the large differences between two dialects [7].

*3.1.8 Amount of data and search space*
Communication with a computer via a microphone creates a large amount of speech data every second. This has to be matched with groups of phones, the sounds, the words and the sentences. Groups of phones build up words and words build up sentences. The number of possible sentences is enormous. The quality of the input, and thereby the amount of input data, can be regulated by the number of samples of the input signal, but the quality of the speech signal will, of course, decrease with a lower sampling rate, resulting in incorrect analysis.

Table 1: Examples of homophones

| the *tail* of a dog | the *tale* of the dog |
|---|---|
| the *sail* of a boat | the *sale* of a boat |

The lexicon, i.e., the set of words, can also be minimized. This introduces another problem, which is called out-of-vocabulary, which means that the intended word is not in the lexicon. An ASR system has to handle out-of-vocabulary words in a robust way [7].

*3.1.9   Ambiguity*
Natural language has an inherent ambiguity, i.e. it is not always easy to decide which of a set of words is actually intended. There are two ambiguities that are particular to ASR, namely, homophones and word boundary ambiguity.

- *Homophones*: The concept of homophones refers to words that sound the same, but have different meanings and different spellings. They are two unrelated words that just happen to sound the same. Some examples of homophones are given in Table 1:
- *Word boundary ambiguity*: When a sequence of groups of phones is put into a sequence of words, the problem of word boundary ambiguity is encountered. Word boundary ambiguity occurs when there are multiple ways of grouping phones into words [7]. An example, taken from [1], illustrates this difficulty:

It's not easy to wreck a nice beach.
It's not easy to recognize speech.
It's not easy to wreck an ice beach.

*3.2 Available Approaches*
This section briefly describes the various techniques that have been used for automating Speech Recognition.

*3.2.1   Dynamic time warping (DTW)-based Speech Recognition*

Dynamic time warping is an approach that was historically used for speech recognition but has now largely been displaced by the more successful HMM-based approach. Dynamic time warping is an algorithm for measuring similarity between two sequences which may vary in time or speed. For instance, similarities in walking patterns would be detected, even if in one video the person was walking slowly and if in another they were walking more quickly, or even if there were accelerations and decelerations during the course of one observation. DTW has been applied to video, audio, and graphics - indeed, any data which can be turned into a linear representation can be analyzed with DTW.

A well known application has been automatic speech recognition, to cope with different speaking speeds. In general, it is a method that allows a computer to find an optimal match between two given sequences (e.g. time series) with certain restrictions, i.e. the sequences are "warped" non-linearly to match each other. This sequence alignment method is often used in the context of hidden Markov models [8].

*3.2.2   Hidden Markov Models*
A hidden Markov model (HMM) is a statistical model in which the system being modeled is assumed to be a Markov process with unknown parameters, and the challenge is to determine the hidden parameters from the observable parameters. The extracted model parameters can then be used to perform further analysis, for example for pattern recognition applications. An HMM can be considered as the simplest dynamic Bayesian network.

In a regular Markov model, the state is directly visible to the observer, and therefore the state transition probabilities are the only parameters. In a *hidden* Markov model, the state is not directly visible, but variables influenced by the state are visible. Each state has a probability distribution over the possible output tokens. Therefore the sequence of tokens generated by an HMM gives some information about the sequence of states.

Hidden Markov models are especially known for their application in temporal pattern recognition such as speech, handwriting, gesture recognition, and bioinformatics.

Modern general-purpose speech recognition systems are generally based on HMMs. These are statistical models which output a sequence of symbols or quantities. One possible reason why HMMs are used in speech recognition is that a speech signal could be viewed as a piecewise stationary signal or a short-time stationary signal. That is, one could assume in a short-time in the range of 10 milliseconds, speech could be approximated as a stationary process. Speech could thus be thought of as a Markov model for many stochastic processes [8].

### 3.2.3 Hybrid Neural Networks

Although the HMM is good at capturing the temporal nature of processes such as speech, it has a very limited capacity for recognizing complex patterns involving more than first order dependencies in the observed data. This is primarily due to the first order state process and the assumption of state conditional observation independence. Neural networks and in particular multi-layer perceptrons (MLPs) are almost the opposite: they cannot model temporal phenomena very well, but are good at recognizing complex patterns in a very parameter efficient way. The Hidden Neural Network hybrid is a very flexible architecture, where the probability parameters of an HMM are replaced by the outputs of small state specific neural networks. One of the models uses the discriminative Conditional Maximum Likelihood (CML) criterion for estimation of the HNN parameters and is normalized globally. The global normalization works at the sequence level and ensures a valid probabilistic interpretation as opposed to the often approximate local normalization enforced in many hybrids. Also, instead of training the HMM and NNs separately, all parameters in the HNN are estimated simultaneously.

HNNs have been applied to speech recognition because of their ability to classify complex patterns along with capturing the temporal nature of speech [8].

### 3.2.4 Neuro-Fuzzy Networks

Fuzzy logic is a form of multi-valued logic derived from fuzzy set theory to deal with reasoning that is approximate rather than precise. Just as in fuzzy set theory the set membership values can range (inclusively) between 0 and 1, in fuzzy logic the degree of truth of a statement can range between 0 and 1 and is not constrained to the two truth values {true, false} as in classic predicate logic. And when *linguistic variables* are used, these degrees may be managed by specific functions.

In the field of artificial intelligence, neuro-fuzzy refers to combinations of artificial neural networks and fuzzy logic. Neuro-fuzzy hybridization results in a hybrid intelligent system that synergizes these two techniques by combining the human-like reasoning style of fuzzy systems with the learning and connectionist structure of neural networks. Neuro-fuzzy hybridization is widely termed as Fuzzy Neural Network (FNN) or Neuro-Fuzzy System (NFS) in the literature. Neuro-fuzzy system (the more popular term) incorporates the human-like reasoning style of fuzzy systems through the use of fuzzy sets and a linguistic model consisting of a set of IF-THEN fuzzy rules. The main strength of neuro-fuzzy systems is that they are universal approximators with the ability to solicit interpretable IF-THEN rules.

The strength of neuro-fuzzy systems involves two contradictory requirements in fuzzy modeling: interpretability versus accuracy. In practice, one of the two properties prevails. The neuro-fuzzy systems in fuzzy modeling research field are divided into two areas: linguistic fuzzy modeling that is focused on interpretability, mainly the Mamdani model; and precise fuzzy modeling that is focused on accuracy, mainly the Takagi-Sugeno-Kang (TSK) model.

The concepts of good interpretability and high accuracy are the requirements for speech recognition as well. However, the tradeoff between them degrades the overall output of the application of neuro-fuzzy techniques to the development of speech recognition technology. Still, the future belongs to neuro-fuzzy computing, especially to the use of these techniques in complex speech [8].

### 3.2.5 Self Organizing Maps

A self-organizing map (SOM) is a type of artificial neural network that is trained using unsupervised learning to produce a low-dimensional (typically two dimensional), discretized representation of the input space of the training samples, called a map. The map seeks to preserve the topological properties of the input space. This makes SOM useful for visualizing low-dimensional views of high-dimensional data. The model was first described as an artificial neural network by the Finnish professor Teuvo Kohonen, and is sometimes called a Kohonen map.

Like most artificial neural networks, SOMs operate in two modes: training and mapping. Training builds the map using input examples. It is a competitive process, also called vector quantization. Mapping automatically classifies a new input vector. Application of SOM's to Speech Recognition is presently arguable [8].

*3.3 Problems Remaining in Speech Recognition as of Today*

Of the problems outlined in Section 3.1, the following problems are central to why some of those problems exist:

- *Differences in accent*: refers to the different ways in which the same word can be spoken,
- *Differences in speed of pronunciation*: refers to various speeds at which words are pronounced,
- *Differences in emphasis*: refers to the different amounts of stress placed on certain words

These three problems are the basis of speaker variability, the difference between spoken language and written language, ambiguity, continuous speech and human comprehension of speech. These are the three problems that have been addressed in Section 4.

## 4. APPLICATION OF FUZZY MATHEMATICS TO SPEECH RECOGNITION OF LINGUISTIC CONTENT

A word in a language is spoken in different accents, different speeds of pronunciation and with different emphasis. For example, the word "vector" of the English language will be spoken by an American as "vektor", with curtness at the 'c' and at the 't', while a Britisher will speak it as "vectorr", with emphasis on the 'c' and a slight repetition on the 'r'. Similarly, a Russian will speak this word as "vecthor", with softness on the 't'.

However, the word remains the same, that is, "vector", with slight variations with respect to different accents, speeds of pronunciation and emphasis. Thus, a single word can be represented as a fuzzy set. However, a word is too specific so as to fit into a generic model of speech recognition. To have a more general model, the fuzzification of phonemes is more appropriate. This model is therefore applied to spoken sentences. One fuzzy set is based on accents, the second on speeds of pronunciation and the third on emphasis. The use of this method will be especially for speech-to-text conversion, by filtering out the unnecessary paralinguistic information from the spoken sentences. This will also lead to the development of a fuzzy filter. To avoid loss of paralinguistic information, which could be useful semantically, the unfiltered signal could be preserved and sent to a paralinguistic information recognition subsystem as well.

The fuzzy set for accent is based on the parameters of softness and sharpness (or curtness) of speaking a phoneme. This has been depicted in Figure 1.

The fuzzy set for the speed of pronunciation is based on the parameters of slow, normal and fast speeds. This has been depicted in Figure 2.

The fuzzy set for emphasis is based on the parameters of light, medium and heavy emphasis

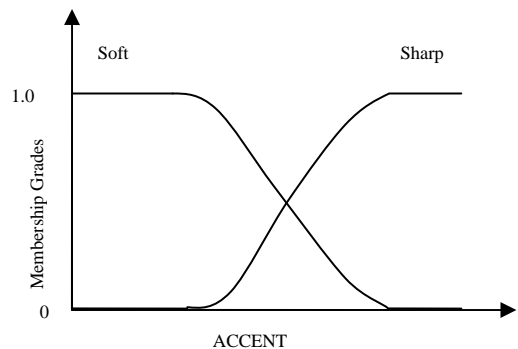
Figure 1: Membership functions for Accent

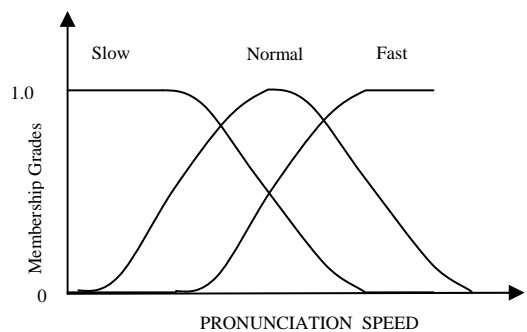
Figure 2: Membership functions for Pronunciation Speed

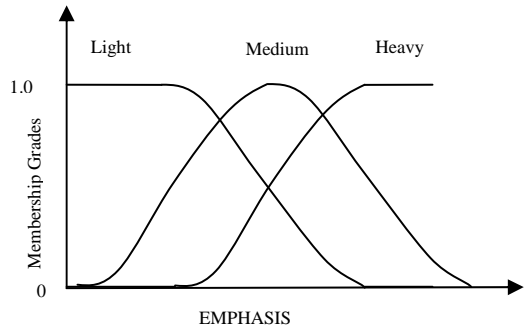

Figure 3: Membership functions for Emphasis

on a phoneme or a group of phonemes. This has been depicted in Figure 3.

Some speakers mispronounce certain words. For these cases, the speech-to-text system must confirm the received words with the speaker, just like a human being does.

Now, combinations of phonemes form words. Since phonemes are to be considered as fuzzy sets as described above, their combinations, namely, words, will be fuzzy sets of fuzzy sets. Further, all linguistic relationships between words can be considered as operations on fuzzy sets. These concepts can be developed further into a fuzzy model of representing grammar if combined with the theory of computation.

## 5. CONCLUSION

Many problems related to ASR have been fully or partially overcome but some problems still remain. To conclude, the techniques of Neuro-Fuzzy or Soft Computing can be applied to implement the concepts presented in this paper to solve the problems of differences in accent, speed of pronunciation and emphasis. The future work related to this paper can be the development of a generic fuzzy model to represent the grammar of a language. Also, the authors are working towards the development of a fuzzy filter based speech-to-text system which will implement the concepts presented in this paper.

## REFERENCES

[1] N. M. Ben Gold, "Speech and Audio Signal Processing, processing and perception of speech and music"; John Wiley & Sons, Inc., 2000.

[2] J. H. M. Daniel Jurafsky, "Speech and Language Processing: An Introduction to Natural Language Processing, Computational Linguistics, and Speech Recognition"; Prentice Hall, New Jersey, USA; 2000.

[3] J. Holmes, "An Introduction to Sociolinguistics"; Longman Group, UK, 1992.

[4] E. A. Jens Allwood, "Corpus-based research on spoken language"; 2001.

[5] K. E. Mats Blomberg, "Automatisk igenkänning av tal"; 1997.

[6] B. Schneiderman, "The limits of speech recognition"; Communications of the ACM, 2000; 43:63–65.

[7] Markus Forsberg, "Why is Speech Recognition Difficult?"; Chalmers University of Technology; February, 2003.

[8] www.wikipedia.org

[9] J.-S.R. Jang, C.-T. Sun, E. Mizutani, "Neuro-Fuzzy and Soft Computing : A Computational Approach to Learning and Machine Intelligence"; Pearson Education, 1997.